%% file: main.tex
\documentclass[lettersize,journal]{IEEEtran}

\usepackage{amsmath,amsfonts,amssymb}
\usepackage{nicefrac}

\usepackage{algorithmic}
\usepackage{algorithm}

\usepackage{graphicx}
\usepackage{array}
\usepackage[caption=false,font=normalsize,
    labelfont=sf,textfont=sf]{subfig}
\usepackage{booktabs}
\usepackage{multirow}
\usepackage{makecell}
\usepackage{stfloats}

\usepackage[T1]{fontenc}
\usepackage{textcomp}
\usepackage{microtype}
\usepackage{xcolor}
\usepackage{pifont}
\usepackage{comment}
\usepackage{listings}
\usepackage{url}
\usepackage{verbatim}

\usepackage{cite}
\usepackage[
    colorlinks=true,
    citecolor=blue,  
    linkcolor=red,   
    urlcolor=blue    
]{hyperref}
\usepackage{cleveref}

\crefname{table}{Table}{Tables}
\Crefname{table}{Table}{Tables}
\crefname{figure}{Fig.}{Figs.}
\Crefname{figure}{Fig.}{Figs.}
\crefname{section}{Section}{Sections}
\Crefname{section}{Section}{Sections}
\crefname{appendix}{Appendix}{Appendices}
\Crefname{appendix}{Appendix}{Appendices}

\begin{document}

\title{PiCo: Physics-Informed Concept Probing for Geo-Foundation Models}

\author{Yihang~She,
        Andrew~Blake,
        Jianbo~Qi,
        and~Srinivasan~Keshav%
\thanks{Yihang She is with the Department of Computer
Science and Technology, University of Cambridge,
Cambridge CB3 0FD, UK
(e-mail: ys611@cam.ac.uk). Corresponding author: Yihang She.}%
\thanks{Andrew Blake is with Clare Hall,
University of Cambridge, Cambridge CB3 9AL, UK
(e-mail: ab@ablake.ai).}%
\thanks{Jianbo Qi is with the Faculty of Geographical
Science, Beijing Normal University,
Beijing 100875, China
(e-mail: qijb@mail.bnu.edu.cn).}%
\thanks{Srinivasan Keshav is with the Department of
Computer Science and Technology, University of
Cambridge, Cambridge CB3 0FD, UK
(e-mail: sk818@cam.ac.uk).}%
\thanks{
Supplementary material is available at this \href{https://drive.google.com/file/d/15eHRDbI04jTTkGBq7Pv555p542fm-Ffi/view?usp=sharing}{link}.}
}

\markboth{Preprint. Under review.}%
{She \MakeLowercase{\textit{et al.}}:
PiCo: Physics-Informed Concept Probing
for Geo-Foundation Models}

%

\maketitle


\begin{abstract}
\input{0_abstract}
\end{abstract}




\input{1_introduction}

\input{2_related_work}

\input{3_methods}

\input{4_results}
\input{5_conclusion}


\section*{Acknowledgment}
This work was supported by the UKRI Centre for Doctoral Training in Application of Artificial Intelligence to the study of Environmental Risks (reference EP/S022961/1) and Cambridge Centre for Carbon Credits. We thank Jovana Knezevic for helpful early discussions on related work in AI interpretability.


\bibliographystyle{IEEEtran}
\bibliography{references}


\end{document}

%% file: 0_abstract.tex
Geo-foundation models (GFMs) are typically evaluated by downstream task predictive performance. This reveals little about which physical properties their embeddings are sensitive to, or which ones are used in predictions.
That limits developers' ability to test hypotheses about model behaviour and users' ability to select models for new tasks.
We propose \emph{Physics-Informed Concept Probing} (PiCo), a post-hoc framework for explaining GFM predictions. 
PiCo identifies concept vectors in a GFM's embedding space by generating
synthetic multispectral satellite observations from a 3D radiative transfer model coupled with a game engine.
These vectors characterise GFM sensitivities to physical properties, informing domain-based hypotheses about downstream performance.
Projecting embeddings onto these concept vectors
produces low-dimensional representations for quantifying concept importance in making predictions.
We evaluate PiCo on three GFMs with different input modalities using seven physical concepts covering forest leaf optics and canopy structure.
In ground-cover estimation, using only RGB values as inputs reduces dry-matter sensitivity and disproportionately degrades predictions for dry vegetation and bare ground, as anticipated by PiCo's sensitivity analysis.
For hierarchical forest type and tree species classification, 
concept representations retain 70.0--93.8\% of full-embedding accuracy.
Different GFMs capture meaningful differences in physical properties between classes, such as canopy structure between broadleaf forest and cleared ground and leaf-water-content between oak and beech. Yet they differ in the strength of these contrasts, and in the concepts they use to drive classification.
PiCo thus provides an interpretable basis for hypothesis testing and model assessment, with potential applicability beyond Earth observation wherever physical models enable controlled simulation and pretrained embeddings are accessible for probing.
Code for PiCo is available at \url{https://github.com/yihshe/PiCo.git}. 

%% file: 1_introduction.tex
\section{Introduction}
Recent advances in representation learning \cite{bengio2013representation} have led to foundation models \cite{bommasani2021opportunities}, which learn general-purpose representations from massive datasets and transfer them across downstream tasks. 
This learning paradigm has seen growing adoption in scientific domains \cite{zhang2024challenges,huang2025survey,ting2025deep} where large-scale observations are available \cite{yang2023medmnist,drusch2012sentinel,majewski2017apache}, including medical imaging \cite{zhang2024challenges}, Earth observation \cite{huang2025survey}, and astronomy \cite{ting2025deep}. 
Earth observation provides a particularly relevant setting, where satellite archives are large, spatially extensive, and repeatedly observed over decades \cite{wulder2022fifty, drusch2012sentinel}.
Recent geo-foundation models (GFMs, $\mathcal{E}$) learn general-purpose representations from satellite observation $\mathbf{x}$ and provide geo-embeddings, $\mathbf{e} = \mathcal{E}(\mathbf{x})$, as reusable features for tasks such as land-cover mapping, crop monitoring, wildfire assessment, and vegetation classification \cite{xiao2025foundation,huang2025survey,lu2025vision}.
Ideally, with precomputed embeddings, users need only train a simple downstream predictor on a small labelled dataset while still achieving high performance.

\begin{figure*}[htbp]
    \centering
    \includegraphics[width=\linewidth]{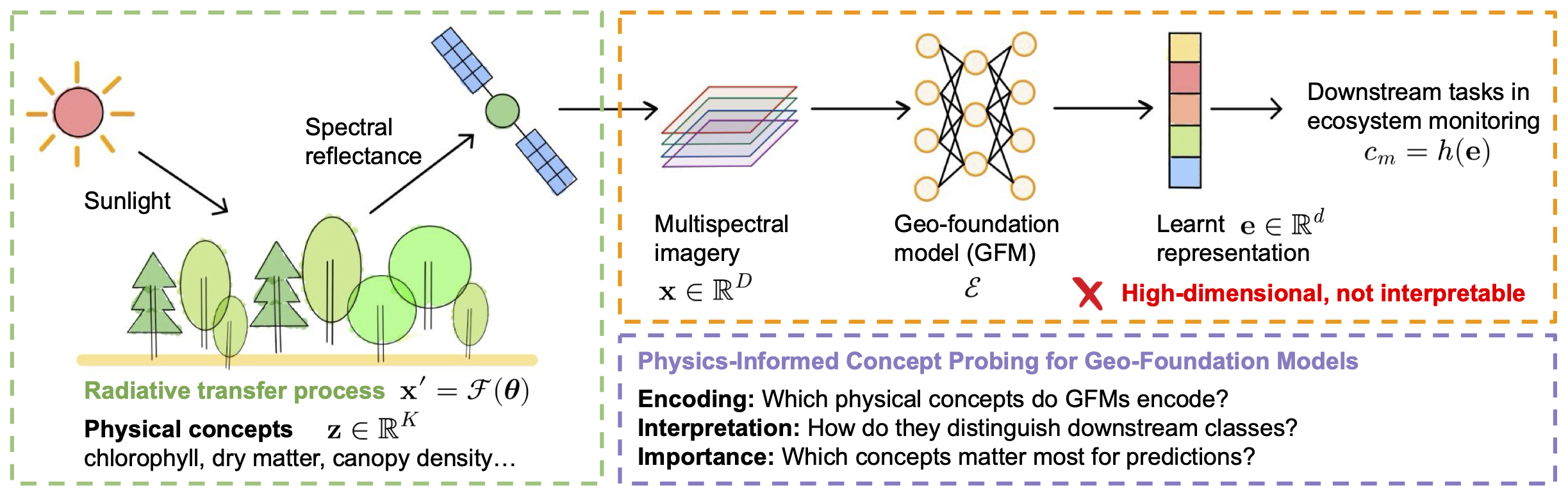}
    \caption{\textbf{PiCo aims to enable physics-grounded interpretation of geo-foundation model (GFM) embeddings by connecting the physical processes underlying satellite observations with learned representations.}
    \textit{
    Standard downstream evaluation maps satellite observations $\mathbf{x}$ to high-dimensional embeddings $\mathbf{e}=\mathcal{E}(\mathbf{x})$ for prediction, without explicitly revealing their physical meaning.
    In contrast, radiative transfer models $\mathcal{F}$ simulate observations $\mathbf{x}'=\mathcal{F}(\boldsymbol{\theta})$ from physically parameterised 3D scenes, where physical concepts $\mathbf{z}$ can be controlled.
    PiCo bridges these two approaches to investigate which physical concepts GFMs encode, how their sensitivities relate to downstream performance, and which concepts matter most for predictions.}
    }
    \label{fig:intro_current_approach}
\end{figure*}

Although using geo-embeddings is attractive for scalability and label efficiency, they are typically high-dimensional representations without explicit semantic or physical meaning \cite{hohl2402opening} (\cref{fig:intro_current_approach}).
Moreover, current benchmarks for evaluating geo-embeddings typically focus on their prediction accuracy \cite{marsocci2024pangaea,lacoste2023geo}.
However, accuracy alone provides limited insight into model behaviour or generalisability.
For example, in a hypothetical task using embeddings to predict wildfires in a forest over 100 days, only one day might have a fire. A model that predicts no fires ever would achieve 99\% accuracy while missing this key event.
A deeper assessment should examine whether the model captures physical features relevant to wildfire risk \cite{yebra2013global}, such as vegetation water and dry-matter content, 
and how these features contribute to its predictions.

More broadly, current evaluation of GFMs leaves two questions unresolved.
First, it gives model developers limited tools to \textit{formulate and test hypotheses} about model performance.
For example, they may want to assess how sensitive a model is to physical features to inform whether it can transfer robustly to wildfire prediction tasks beyond benchmark settings.
Second, it gives downstream users limited ability to \textit{build trust}: with more than 150 GFMs now available \cite{xiao2025foundation,huang2025survey,lu2025vision} --- each trained with different modalities and learning objectives --- knowing how physical features of interest contribute to predictions would allow users to apply their domain knowledge in choosing a model.

Geo-embeddings are learned from Earth observations, which are essentially measurements generated from underlying physical states through physical and instrumental processes.
Mechanistic forward models $\mathcal{F}$ describe these processes and map physical states $\boldsymbol{\theta}$ to modelled observations, $\mathbf{x}'=\mathcal{F}(\boldsymbol{\theta})$ \cite{lanconelli2025fifth} (\cref{fig:intro_current_approach}).
In optical remote sensing, radiative transfer models (RTMs) links properties such as leaf water and dry-matter content to observed spectral signals \cite{schaepman2006reflectance}.
We ask whether these physical processes can be connected to learned representations to improve physical interpretability while retaining the scalability of GFMs.

Prior work in computer vision identifies interpretable directions in embedding spaces, known as \textit{concept vectors}, from image datasets annotated with human-defined concepts, and uses these vectors to explain model predictions \cite{kim2018interpretability,zhou2018interpretable,yeh2020completeness}.
However, in Earth observation, annotations for physical variables often require extensive field or laboratory measurements \cite{kira2015non,shen2020quantifying}.
Even where measurements exist, real-world observations rarely provide controlled variation of individual physical variables, making their effects difficult to isolate \cite{zhan2026inferring}.
Similar challenges arise across scientific domains \cite{cranmer2020frontier}, and have motivated the use of simulations of forward models to infer physical variables underlying observations \cite{tarantola2005inverse,lanconelli2025fifth,garcia2016aspcap}. 
In this study, we use controlled simulations to vary these variables systematically and identify concept vectors from the resulting changes in the embedding space.

We present PiCo --- Physics-Informed Concept Probing, a general post-hoc explainability framework for scientific foundation models, demonstrated here for GFMs. 
It requires only a forward physical model, the ability to run controlled simulations of observations, and an embedding space constructed from those observations.
With GFMs as our testbed, we treat multispectral satellite observations as the outputs of a physical rendering process and use a recent 3D radiative transfer model (RTM) \cite{qi2019less, qi2026retrieving} as the corresponding graphics renderer.
We extend CAMP3D \cite{she2025scaling}, a game-engine LiDAR data pipeline for forest vision tasks, with this 3D RTM rendering capability.
This allows us to generate a comprehensive synthetic dataset of Sentinel-2-like observations from diverse, structurally realistic forest scenes, with controlled variations in physical concepts covering leaf optics and canopy structure.

Pairing ground-truth physical variables with embeddings from synthetic observations allows us to learn concept vectors that capture concept-sensitive directions in a GFM embedding space using simple linear probes.
Once learned, the concept vectors provide a low-dimensional, interpretable basis for otherwise opaque embeddings.
For a given embedding, we construct its concept representation from its standardised projections onto these vectors.
We then use this representation as a surrogate for the original embedding in downstream tasks and apply SHAP \cite{lundberg2017unified} to derive concept importance scores, quantifying the global importance of each physical concept for predicting each class.
Knowing which physical concepts GFMs are sensitive to also helps us formulate hypotheses about their performance using domain knowledge.

We apply PiCo to three GFMs with different input modalities: the RGB-based DINOv3-Sat \cite{simeoni2025dinov3} and the multispectral Prithvi-v2 \cite{szwarcman2025prithvi} and OlmoEarth \cite{herzog2025olmoearth}.
Using the SLATS ground-cover dataset \cite{muir2011factional}, we test whether sensitivity to physical concepts helps anticipate performance when estimating the fractions of cover types with distinct physical properties.
We also use TreeSatAI \cite{ahlswede2022treesatai} to quantify concept importance in downstream classification at both the forest-cover and fine-grained species levels.

Our experiments show that input modality affects sensitivity to specific physical concepts.
RGB-only inputs limit sensitivity to leaf dry-matter content, motivating the hypothesis that these models will struggle to distinguish dry vegetation from bare ground.
The experiments support this hypothesis: the RGB-only restriction disproportionately degrades fractional-cover estimation for dry vegetation and bare ground compared to green vegetation.
For tree classification, the concept representations retain 70.0--93.8\% of full-embedding accuracy.
Models capture similar physically meaningful contrasts, such as the low leaf area index and vegetation cover of cleared areas, yet their importance scores reveal that they rely on different concepts for classification.
Alternatively, the original embeddings can be retained for classification alongside the concept representations for interpretation, preserving classification accuracy but providing approximate explanations. 
These findings provide a basis for GFM developers to validate and refine model design through physical sensitivity tests, while importance scores derived from sensitive concepts offer greater transparency to practitioners and domain experts.

To summarise, our contributions are as follows.
\begin{itemize}
    \item We introduce PiCo, a post-hoc, physics-informed explainability framework for geo-foundation models. PiCo learns concept vectors from physical simulations to support hypothesis testing about model performance and explain downstream predictions through physical concepts.
    
    \item We extend CAMP3D with 3D radiative transfer modelling to create a synthetic data pipeline that generates controlled Sentinel-2-like observations from game-engine forest scenes, with ground-truth physical variables for identifying concept vectors.
    
    \item We evaluate PiCo on three geo-foundation models and real-world downstream datasets, demonstrating how physical sensitivities inform testable hypotheses about model performance and how concept importance reveals differences in the physical concepts used for classification.
\end{itemize}

We have not validated PiCo in other scientific domains; however, it may be useful in domains where similar conditions hold, i.e.: a physical model exists, observation can be simulated, and an embedding space is accessible for probing.

%% file: 2_related_work.tex
\section{Related Work}
\subsection{Geo-foundation models and current evaluations}
Geo-foundation models (GFMs) are based on representation learning, which learns general-purpose features from data that can be used by simple predictors for various downstream tasks \cite{bengio2013representation}.
The success of data-driven representation learning and the abundance of Earth observation, has led to the development of more than 150 GFMs \cite{huang2025survey,lu2025vision}, with more emerging.
The GFMs considered in this study have distinct observational modalities.
DINOv3-Sat \cite{simeoni2025dinov3} extends a popular general-vision training recipe to RGB satellite imagery.
Prithvi-v2 \cite{szwarcman2025prithvi} is trained on additional optical bands in the near-infrared and short-wave infrared ranges, in addition to RGB bands.
OlmoEarth \cite{herzog2025olmoearth} is trained on both passive optical and active microwave bands, which capture surface-structure changes, and incorporates supervision from products such as land-cover and canopy-height maps.
Temporality is another key aspect of Earth observation, and several models focus on generating annual embeddings from time series of Earth observations \cite{brown2025alphaearth,feng2026tessera}.
Here, we focus on single-timestamp satellite imagery, as simulating temporal dynamics introduces additional complexity beyond the scope of this study.

The development of current GFMs remains predominantly data-driven \cite{xiao2025foundation}, while benchmarks largely focus on the predictive accuracy of representations learned under this paradigm \cite{marsocci2024pangaea,lacoste2023geo}.
As a result, GFM development makes limited use of the physical mechanisms behind Earth observations. 
Validation metrics alone do not make model behaviour explainable \cite{hohl2402opening}; and without explainability, users often find it difficult to choose among many models \cite{hanna2026geosane}.
\cite{rahman2026physically} analyse the physical implications of AlphaEarth's 64-dimensional embeddings by relating individual embedding dimensions and 26 environmental variables, finding some features strongly linked to specific variables.
However, these correlations do not establish how the environmental variables relate to downstream predictions made from the embeddings or quantify their importance to those predictions.
Such dimension-wise analysis also becomes harder to interpret for higher-dimensional geo-embeddings, where semantic information may be distributed across multiple, potentially redundant dimensions.

\subsection{Concept vectors and interpretability in representation learning}

The idea that concept directions exist in representations learned by neural networks can be traced back to work in natural language processing \cite{mikolov2013efficient}.
A famous example is that subtracting ``man'' from ``king'' and adding ``woman'' yields an embedding close to ``queen'', indicating a gender-related direction in the space.
\cite{alain2016understanding} showed that information becomes more linearly accessible in deeper network layers, as measured using linear probes.
Subsequent works extended this idea by explicitly identifying \emph{concept vectors}, which are usually defined as unit vectors representing linear directions associated with concepts in neural representation spaces \cite{kim2018interpretability,zhou2018interpretable,yeh2020completeness}.
In vision domains, they are often learned from image datasets annotated with human-defined concepts \cite{kim2018interpretability,zhou2018interpretable}.
Concept vectors can also emerge through learning, as in ConceptSHAP \cite{yeh2020completeness}.
This is related to disentangled representation learning, which seeks low-dimensional, interpretable representations within high-dimensional features \cite{wang2024disentangled,eastwood2022dci}, although the learned factors are not always physically meaningful.
Rather than relying on potentially arbitrary human-defined concepts or annotations that are often scarce for physical properties in Earth observation, PiCo uses controlled simulations from predefined 3D radiative transfer models to identify physical concept vectors in GFM representations.

To relate concept vectors to model predictions, TCAV \cite{kim2018interpretability} computes first-order directional derivatives that measure how a model's score for a target class changes when its representation is perturbed along a concept vector.
However, these directional derivatives do not directly quantify the relative contribution of each concept to a prediction when multiple concepts are considered together.
TCAV also requires gradients that propagate through the downstream predictor, making it not directly applicable to non-parametric probes like $k$NN.
Other approaches attribute concept importance by accounting for nonlinear interactions among concepts \cite{zhou2018interpretable,yeh2020completeness}.
In particular, SHAP \cite{lundberg2017unified} has been used as a model-agnostic approach to evaluate a concept's marginal contribution across subsets of other concepts \cite{yeh2020completeness} using Shapley values from cooperative game theory \cite{shapley1953value}, and is used in this study. 
Like other concept-based explanation frameworks \cite{kim2018interpretability,zhou2018interpretable}, PiCo is post-hoc: it analyses trained models without modifying their architecture or retraining their parameters, making it applicable to existing geo-foundation models.
Alternative approaches like Concept Bottleneck Models \cite{koh2020concept} create explicitly interpretable layers during training, which most existing GFMs lack.

\subsection{Probing representations with synthetic data from controlled simulations}

Synthetic data from graphics renderers have been used to evaluate properties such as disentanglement and generalisation in learned representations \cite{dsprites17,3dshapes18,moreno2016overcoming,reed2015deep,eastwood2022dci,schott2021visual}.
Compared with real data, synthetic data can be generated from controlled simulations, allowing individual factors to be isolated and ground-truth generative variables to be made available for evaluation.
Existing datasets such as dSprites \cite{dsprites17}, Shapes3D \cite{3dshapes18}, and Cars3D \cite{reed2015deep} typically consist of 2D images of a single object, generated by manipulating factors such as pose, size, and lighting.
Optical sensing in Earth observation can also be viewed as a graphics rendering process \cite{she2025pila}, and 3D radiative transfer models that simulate this process can be viewed as graphics renderers.
Three-dimensional RTMs \cite{Eradiate,qi2019less,wang20243d,qi2026retrieving} model the formation of optical satellite observations by simulating how radiation interacts with vegetation structure, leaf optical properties, soil, and illumination-viewing geometry.
In particular, the latest LESS 3D RTM \cite{qi2026retrieving} is built on the differentiable Mitsuba 3 graphics-rendering framework \cite{jakob2022mitsuba3}.
However, compared to the synthetic vision datasets above, rendered satellite imagery differs substantially in image formation, especially in vegetation biophysical processes on Earth’s surface \cite{feret2017prospect}, as well as in spectral range, scale, and resolution \cite{qi2019less, drusch2012sentinel}.

Previous work has rendered both 2D images and 3D LiDAR point clouds of vegetation from graphics engines for ground-based forest vision tasks \cite{feng2025spread,raistrick2023infinite,she2025scaling}.
We extend the CAMP3D pipeline of \cite{she2025scaling}, which renders LiDAR point clouds from game-engine forest scenes, to also render satellite images by replacing the LiDAR simulator \cite{winiwarter2022virtual} with a 3D RTM \cite{qi2019less}. 
Learning concept vectors from synthetic data is also related to simulation-based inference \cite{cranmer2020frontier} --- in remote sensing, simulations from a forward RTM have been used to learn an inverse mapping from satellite observation to biophysical variables \cite{weisss2toolbox,she2025pila,qi2026retrieving}.

%% file: 3_methods.tex
\section{Methods}

\subsection{Problem setup}
\begin{figure*}[htbp]
    \centering
    \includegraphics[width=\linewidth]{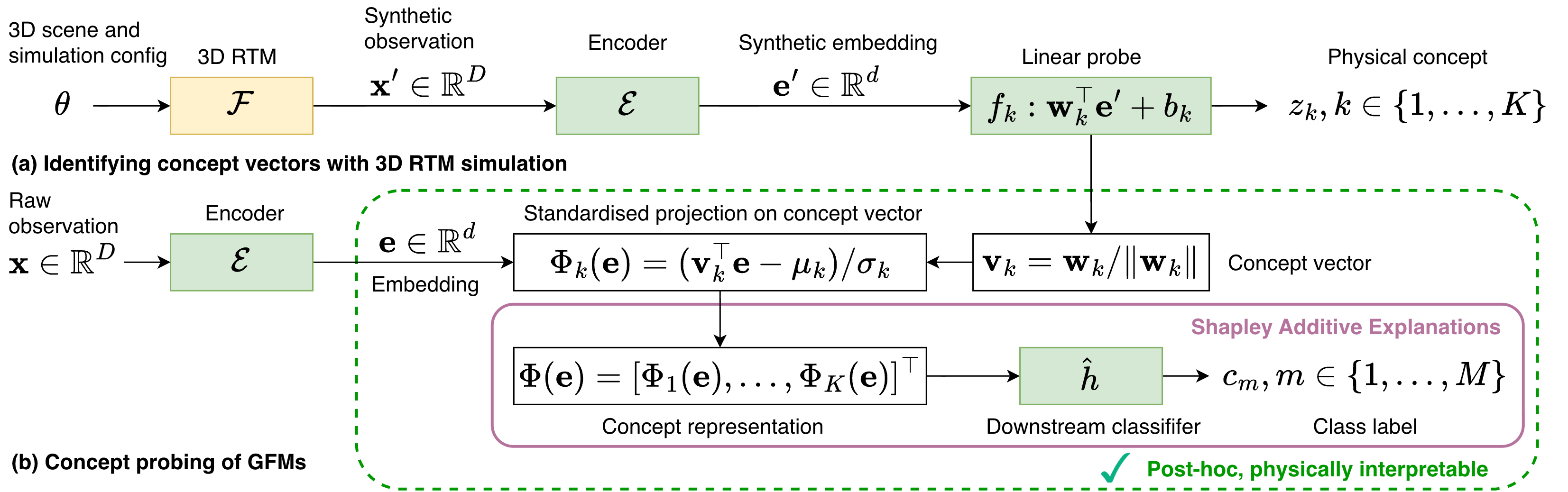}
    \caption{\textbf{Physics-informed concept probing for geo-foundation models.}
    \textit{
    (a) We build a physical simulation pipeline in which simulator configurations including 3D scenes $\boldsymbol{\theta}$ are rendered by a 3D radiative transfer model $\mathcal{F}$ into synthetic observations $\mathbf{x}'=\mathcal{F}(\boldsymbol{\theta})$, while the same configurations provide physical concept variables $\mathbf{z}=g(\boldsymbol{\theta})$. For each physical concept variable $z_k,k=\{1,\dots,K\}$, a linear probe $f_k(\mathbf{e})=\mathbf{w}_k^\top \mathbf{e}+b_k$ is trained on synthetic embeddings paired with ground truth in the form of continuous physical variables $(\mathbf{e',z_k})$. Its normalised weight vector $\mathbf{v}_k=\mathbf{w}_k/\|\mathbf{w}_k\|$ defines the concept vector.
    (b) For a real-world observation $\mathbf{x}$, its embedding $\mathbf{e}$ is projected onto the learned concept vectors. The projection for each concept is standardised to place all concepts on a comparable scale, and the resulting values are stacked to form the concept representation $\Phi(e)\in\mathbb R^K$. We then evaluate whether this representation preserves downstream decision capability and use Shapley Additive Explanations (SHAP) to quantify a geo-foundation model's reliance on each physical concept.
    }
    }
    \label{fig:methods_diagram}
\end{figure*}

Consider forest-type classification as an illustrative example. Given multispectral satellite observations of forest plots, a $k$NN classifier can be trained on geo-foundation model embeddings to predict forest types, such as broadleaf or coniferous. Beyond classification accuracy, we seek to understand how these predictions relate to physical concepts, such as leaf chlorophyll content and fractional vegetation cover. PiCo provides a framework for investigating these relationships, as illustrated in \cref{fig:methods_diagram}.

A 3D radiative transfer model (RTM) takes a configuration $\boldsymbol{\theta}$ comprising 3D scene geometry, surface optical properties, and illumination and viewing conditions \cite{qi2019less}. Surface optical properties specify wavelength-dependent reflectance and transmittance, while scene and acquisition geometry determine the ray-tracing interactions. The RTM $\mathcal{F}$ maps this configuration to a synthetic multispectral observation:
\begin{equation}
    \mathbf{x}' = \mathcal F(\boldsymbol{\theta}) \in \mathbb R^D,
\end{equation}

From the same simulator configuration $\boldsymbol{\theta}$, we obtain the values of selected physical concepts of interest,
\begin{equation}
    \mathbf{z} = g(\boldsymbol{\theta}) \in \mathbb{R}^K,
\end{equation}
where $\mathbf{z} = [z_1,\dots,z_K]^\top \in \mathbb R^K$ denotes the values of the $K$ physical concepts considered in the study. 
In our experiments, which focus on forest vegetation, these concepts are leaf-optical such as chlorophyll content or structural such as canopy density. 

While the 3D RTM provides a controlled physical simulation with known values of the concepts of interest, our ultimate goal is to interpret model behaviour on real-world downstream tasks.
We therefore consider a downstream dataset containing real satellite observations $\mathbf{x}$ and class labels $\mathbf{c}=[c_1,\dots,c_M]^\top\in\mathbb{R}^M$, where $M$ is the number of classes. 
A geo-foundation model encoder $\mathcal E$ maps each observation to an embedding,
\begin{equation}
    \mathbf{e} = \mathcal E(\mathbf{x}) \in \mathbb R^d .
\end{equation}

A downstream predictor can then be applied to the embeddings to predict the class labels,
\begin{equation}
     h(\mathbf{e}):\mathbb{R}^{d}\rightarrow[0,1]^{M},
\end{equation}
where $h$ is a multi-class classifier. Assuming an embedding that is effective, the classifier $h$ often need not be very complex --- it could be something as simple as a $k$-nearest neighbours ($k$NN) \cite{herzog2025olmoearth} classifier.

\subsection{Physics-informed concept probing}
\paragraph{Physically controlled simulation pipeline.}
To learn concept vectors corresponding to the selected physical concepts, we construct a physically controlled data-generation pipeline by combining diverse, structurally realistic 3D forest scenes sourced from game engines \cite{she2025scaling} with the LESS 3D radiative transfer model \cite{qi2019less}. 
The 3D RTM acts as a graphics renderer incorporating biophysical processes, generating multispectral observations $\mathbf{x}'$ paired with corresponding values of physical concepts $\mathbf{z}$. The concepts studied in our experiments are described in \cref{sec:eval_protocol}, with further simulation details provided in Section S1 of the Supplementary Material.

\paragraph{Learning concept vectors from physical simulation.}  
Let $z_k$, for $k\in\{1,\dots,K\}$, denote the value of the $k$-th physical concept. We compute embeddings from synthetic observations $\mathbf{e'} = \mathcal{E}(\mathbf{x'})$. 
For each physical concept, we fit a linear probe to predict the calibrated physical concept value from synthetic embeddings (\cref{fig:methods_diagram} (a)): 
\begin{equation}
    f_k(\mathbf{e'})=\mathbf{w}_k^{\top}\mathbf{e'}+b_k.
\end{equation}
Following previous work on linear probing of neural representations \cite{kim2018interpretability}, we use ridge regression, whose L2 regularisation on probe weights helps avoid overfitting on certain features in its estimations when high-dimensional embeddings often have redundant or correlated features, by solving
\begin{equation}
    (\mathbf{w}_k,b_k) = \underset{\mathbf{w},b}{\arg\min}\;\frac{1}{N'}\sum_{n=1}^{N'}\left(z_{n,k} -\mathbf{w}^{\top}\mathbf{e}'_n - b \right)^2 + \lambda\lVert\mathbf{w}\rVert_2^2,
    \label{eq:ridge_concept_probe}
\end{equation}
where $N'$ is the number of samples of the synthetic dataset. 

The weight vector $\mathbf{w}_k$ defines the direction in embedding space along which along which the representation is most linearly sensitive to that concept. 
We therefore define the \textit{unit concept vector}:
\begin{equation}
    \mathbf{v}_k = \frac{\mathbf{w}_k}{\lVert\mathbf{w}_k\rVert} \in \mathbb{R}^d.
    \label{eq:concept_direction}
\end{equation}
The orientation of $\mathbf{v}_k$ is determined by the regression target, so movement in its positive direction corresponds to an increase in the probe prediction for $z_k$, and vice versa.

The affine prediction $f_k(\mathbf{e}')$ combines a sample's position along $\mathbf{v}_k$ with a scale and offset learned from the synthetic data, represented by $\lVert\mathbf{w}_k\rVert_2$ and $b_k$.
Our analysis focuses on the former: whether variation along $\mathbf{v}_k$ distinguishes classes in a downstream task.
We do not assume that the learned scale and offset transfer unchanged to real downstream scenes, since differences between synthetic and real scene distributions may affect the appropriate scale and offset. These parameters are not required for our analysis.

\paragraph{Concept representation.} 
For each downstream task, given embeddings $\mathbf{e}$ extracted from task-specific regions of interest, we project them onto the concept vector $\mathbf{v}_k$ and centre and scale $\mathbf{v}_k^\top\mathbf{e}$ to obtain standardised projections:
\begin{equation}
    \Phi_k(\mathbf{e}) = \frac{\mathbf{v}_k^\top\mathbf{e}-\mu_k}{\sigma_k},
    \label{eq:standardised_concept_coordinate}
\end{equation}
where $\mu_k$ and $\sigma_k$ are, respectively, the mean and standard deviation of raw projections over the training samples of the downstream task. Centring removes the dependence of $\Phi_k(\mathbf{e})$ on any global shift in the origin of the embedding space. Scaling preserves the relative ordering of samples along the concept vector $\mathbf{v}_k$ while making $\Phi_k(\mathbf{e})$ on different concept vectors comparable under the same scale.
The \textit{concept representation} of an embedding is defined as the concatenation of its standardised projections onto individual concept vectors:
\begin{equation}
    \Phi(\mathbf{e}) = \begin{bmatrix}\Phi_1(\mathbf{e}),\dots,\Phi_K(\mathbf{e})\end{bmatrix}^{\top}\in\mathbb{R}^{K}.
    \label{eq:concept_projected_representation}
\end{equation}

\begin{figure}[htbp]
    \centering
    \includegraphics[width=\linewidth]{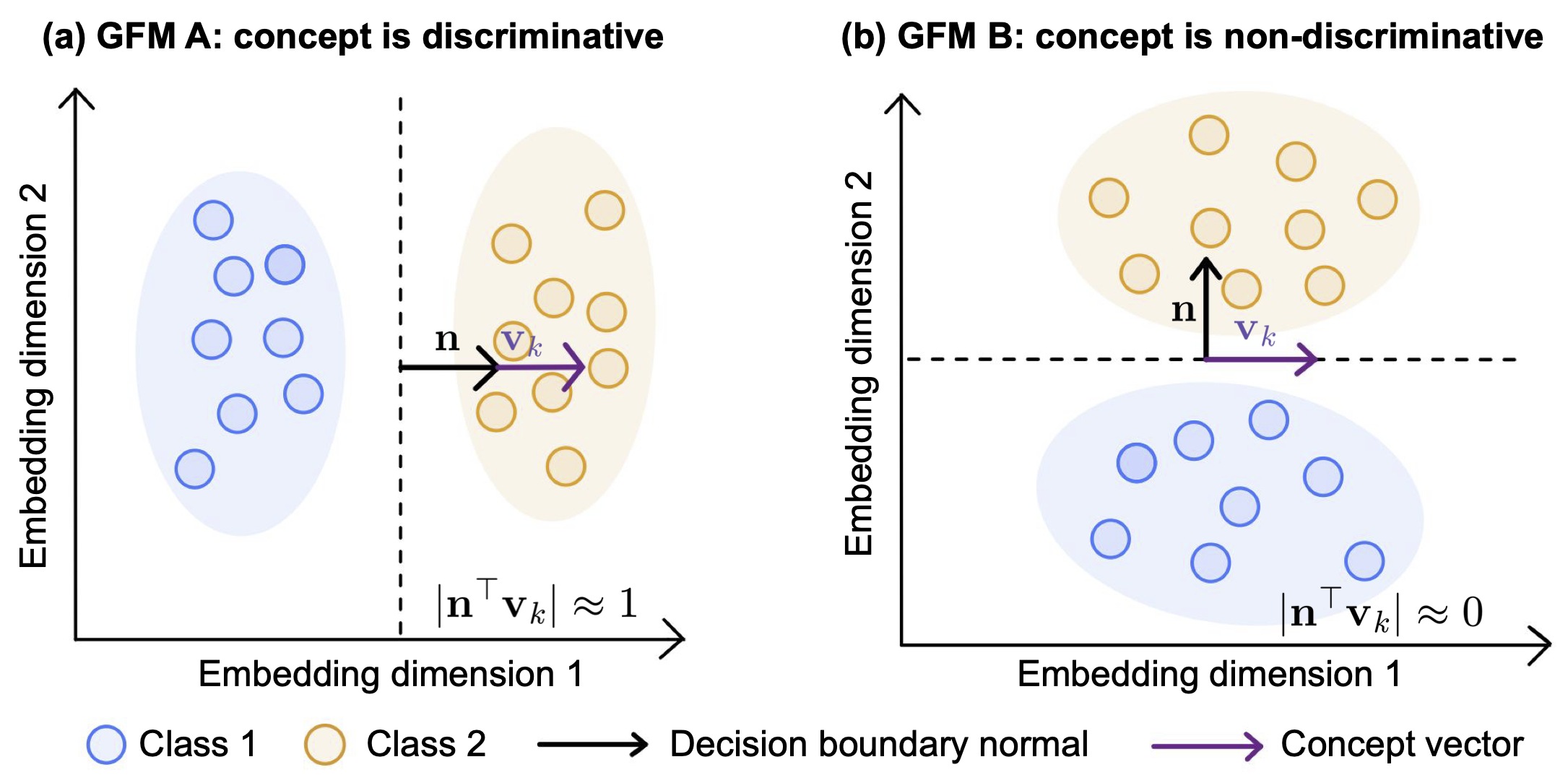}
    \caption{\textbf{The discriminative relevance of the same physical concept varies across GFMs due to differences in their learned embedding geometries,} demonstrated here with a simple 2D example.
    \textit{
    The two classes $c_1$ and $c_2$ are separated by a linear decision boundary (dashed line), with unit normal $\mathbf{n}$ (black arrow) and concept vector $\mathbf{v}_k$ (purple arrow).
    (a) In the embedding space of GFM A, $\mathbf{v}_k$ aligns with $\mathbf{n}$, so the concept projections separate the two classes.
    (b) In the embedding space of GFM B, $\mathbf{v}_k$ is orthogonal to $\mathbf{n}$, so the concept projections overlap, providing little class discrimination.
    }
    }
    \label{fig:methods_concept_vector_discrimination}
\end{figure}

The relevance of a concept vector $\mathbf{v}_k$ to a downstream classification task can also be understood geometrically (\cref{fig:methods_concept_vector_discrimination}). Consider a linear hyperplane that best separates two classes $c
_1$ and $c_2$ in the embedding space, and let $\mathbf{n}$ denote its unit normal. $\mathbf{n}$ represents the linear direction where the projections of the two class clusters are best separated. 
The magnitude of
$\mathbf{n}^{\top}\mathbf{v}_k$
measures the alignment between the hyperplane normal $\mathbf{n}$ and the concept vector $\mathbf{v}_k$. 
When the two directions are aligned, there will be strong discrimination between $\Phi_k(\mathbf{e})~\vert~c_1$ and $\Phi_k(\mathbf{e})~\vert~c_2$. 
When they are orthogonal, $\mathbf{v}_k$ is parallel to the hyperplane and $\Phi_k(\mathbf{e})$ provides little discrimination between these two classes. 
Each geo-foundation model induces its own embedding geometry, so the relative arrangement of class clusters with respect to $\mathbf{v}_k$ may vary across models. As a result, a concept that is strongly associated with a downstream class in one embedding space may be weakly associated with the same class in another.

\paragraph{Concept importance attribution.}
To quantify the contributions of physical concepts to class prediction scores, we fit a second classifier that uses only the concept representation $\Phi(\mathbf{e})$:
\begin{equation}
    \hat{h}(\Phi(\mathbf{e})):\mathbb{R}^{K}\rightarrow[0,1]^{M}.
\end{equation}
The classifier $\hat{h}$ uses the same classifier family as $h$, which operates on the original embedding $\mathbf{e}$, but is fitted, instead, on $\Phi(\mathbf{e})$. Both classifiers use the same training samples and ground-truth downstream class labels; they differ in their input representations.
To assess the predictive sufficiency of the concept representation, we primarily compare the accuracy of $\hat{h}$ with that of $h$ on the same held-out samples, reporting the proportion of full-embedding accuracy retained by the concept-based classifier.

The geometric intuition discussed earlier considers binary class separation along a single concept vector. Often, the downstream tasks contains multiple classes, and the concept-space classifier $\hat{h}$ uses all concept projections $\Phi(\mathbf{e})$ jointly. A concept may show limited class separation in isolation but still substantially affect the probability assigned to a class when considered together with other concepts. 
We therefore use Shapley Additive Explanations (SHAP) \cite{lundberg2017unified}, an attribution
framework based on Shapley values \cite{shapley1953value}. For each sample and class probability, SHAP quantifies the contribution of a concept by averaging its marginal effect over all subsets of the remaining concepts. 
Let
\begin{equation}
    \Phi^{(i)} = \Phi(\mathbf{e}^{(i)})
\end{equation}
denote the concept representation of the geo-embedding for $\mathbf{e}^{(i)}$ for the $i$-th test sample, where $i \in\{1,\dots,N\}$. Let 
\begin{equation}
    F=\{1,\ldots,K\}
\end{equation}
denote the indices of the $K$ physical concepts used for attribution. 
Given a test sample represented by $\Phi^{(i)}$ and $\sum_{m=1}^{M}\hat{p}_m = 1$, the multi-class classifier $\hat{h}$ predicts a distribution of probability on the labels of the class $\mathbf{c}=[c_1,\dots,c_M]^\top\in\mathbb{R}^M$: $\mathbf{\hat{p}} = [\hat{p}_1,\dots,\hat{p_M}]^\top\in\mathbb{R}^M$, where $\hat{p}_m$ denotes the probability assigned to $c_m$ by $\hat{h}$. 

For a subset $S\subseteq F$, we estimate the probability assigned to class $c_m$ when only the concept values indexed by $S$ are retained from $\Phi^{(i)}$, while the remaining concepts are marginalised with respect to a background distribution.
To do so, we construct masked samples by retaining the concept values of $\Phi^{(i)}$ indexed by $S$ and replacing the remaining values with those from background samples drawn from the training set used to fit $\hat{h}$.
We denote by $\hat{p}_m(S;\Phi^{(i)})$ the probability assigned to class $c_m$ by the fixed classifier $\hat{h}$, averaged over these background replacements.

For evaluation sample $\Phi^{(i)}$, concept $z_k$, class $c_m$, the Shapley value \cite{shapley1953value,lundberg2017unified} is

\begin{equation}
\begin{aligned}
\psi_{i,k,m}
&= \sum_{S\subseteq F\setminus\{k\}}
\frac{|S|!(K-|S|-1)!}{K!} \\
&\quad \times
\left[
\hat{p}_m(S\cup\{k\};\Phi^{(i)})
-
\hat{p}_m(S;\Phi^{(i)})
\right].
\end{aligned}
\label{eq:concept_shap}
\end{equation}

The term in brackets is the marginal change in the probability $\hat{p}_m$ assigned to class $c_m$, when concept $z_k$ is added to the subset of concepts indexed by $S$ for sample $\Phi^{(i)}$. 
Because the number of subsets varies with subset size $\vert S \vert$, the factorial weight assigns equal total weight, $1/K$, to each subset size, distributed equally among all subsets of that size.
A positive sign of $\psi_{i,k,m}$ indicates that adding concept $z_k$ increases the probability assigned to $c_m$ for sample $\Phi^{(i)}$; and vice versa. Its magnitude indicates the strength of this contribution, irrespective of its direction. 
For $\Phi^{(i)}\in\mathbb{R}^K$, whose dimensionality is assumed to be much smaller than that of $\mathbf{e}\in\mathbb{R}^d$, we explicitly evaluate all $2^K$ possible concept subsets.

To obtain a dataset-level score of concept importance for predicting each class, we average the absolute values $\psi_{i,k,m}$ over the test samples $\Phi^{(i)}, i=\{1,\dots,N\}$:
\begin{equation}
    A_{k,m} = \frac{1}{N} \sum_{i=1}^{N}\left|\psi_{i,k,m}\right|.
    \label{eq:raw_concept_importance}
\end{equation}
Taking the absolute value prevents positive and negative contributions from cancelling when aggregated across samples \cite{eastwood2022dci}. Thus, $A_{k,m}$ measures how strongly concept $z_k$ affects the probability assigned to class $c_m$ across the test set. 

We then normalise the importance values in $\mathbf{A}\in\mathbb{R}^{K\times M}$ separately for each class to obtain an importance score for each concept-class pair in a downstream task:
\begin{equation}
    R_{k,m} = \frac{A_{k,m}}{\sum_{k=1}^{K}A_{k,m}}.
    \label{eq:normalised_concept_importance}
\end{equation}

Therefore, each column of $\mathbf{R}\in\mathbb{R}^{K\times M}$ sums to one, and
$R_{k,m}$ represents the relative importance of concept $z_k$ for predicting class $c_m$.

\section{Experimental protocol} \label{sec:eval_protocol}
\paragraph{Physical concepts and simulation settings.}

We study seven physical concepts derived from the simulator configuration \(\boldsymbol{\theta}\), comprising three leaf-optical variables and four structural variables. The leaf-optical variables are key parameters governing leaf reflectance and transmittance of the leaf surface \cite{feret2017prospect}: \emph{chlorophyll content}, \emph{dry matter}, and \emph{water content}. These variables represent important leaf-level functional traits and have distinct spectral signatures in Sentinel-2 observations. Chlorophyll primarily affects absorption in the visible wavelengths, while dry matter is characterised mainly by absorption in the short-wave infrared (SWIR), with contrast against the near-infrared region. Water content is likewise expressed through SWIR absorption, with Sentinel-2 band B11 providing sensitivity to liquid water.
We additionally derive four structural concepts from the 3D scene geometry that capture key aspects of forest canopy structure \cite{qi2019less}: \emph{canopy height}, \emph{fractional vegetation cover}, \emph{leaf area index}, and \emph{canopy density}. 

\begin{figure*}[htbp]
    \centering
    \includegraphics[width=\textwidth]{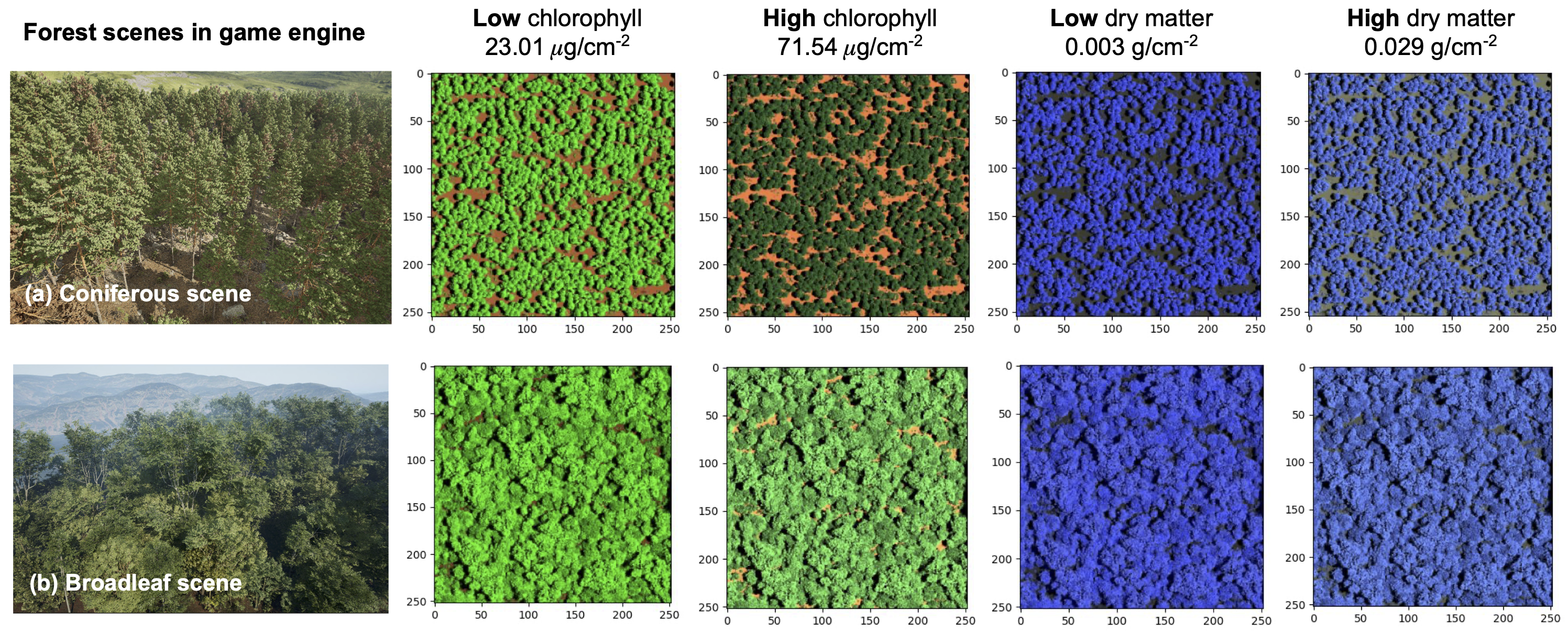}
    \caption{\textbf{PiCo integrates 3D radiative transfer modelling with structurally realistic forest scenes from game engine to generate synthetic multispectral observations under controlled biophysical variations.}
    \textit{
    Examples show coniferous (top) and broadleaf (bottom) forest scenes, alongside simulated satellite imagery at 1 m resolution with varying leaf chlorophyll and dry matter contents.
    Chlorophyll variations are visualised using RGB composites, where higher chlorophyll content increases red-band absorption.
    Dry matter variations are visualised using false-colour composites (B12, B11, B8A), where higher dry matter content generally increases shortwave infrared (SWIR) absorption.
    The same leaf optical parameters are applied to both forest scenes within each column.
    }}
    \label{fig:data_camp3d_rtm_simulation}
\end{figure*}

In total we simulated 21,000 synthetic satellite images from 10 structurally realistic, diverse forest scenes (4 coniferous, 6 broadleaf) built using the CAMP3D tool \cite{she2025scaling} (\cref{fig:data_camp3d_rtm_simulation}). \cref{tab:semiless_prospect_params} summarizes the simulation configurations, with details in Section S1 of the Supplementary Material.

\begin{table}[htbp]
\centering
\caption{\textbf{Leaf optical variables and observation geometry
used in LESS 3D radiative transfer simulations.}
Num denotes the number of sampled values;
LHS denotes Latin hypercube sampling.}
\label{tab:semiless_prospect_params}

\footnotesize
\setlength{\tabcolsep}{2.5pt}
\renewcommand{\arraystretch}{1.15}

\begin{tabular}{@{}llccccc@{}}
\toprule
 & \textbf{Variable} & \textbf{Unit}
 & \textbf{Min} & \textbf{Max}
 & \textbf{Step} & \textbf{Num} \\
\midrule

\multirow{6}{*}{\makecell[l]{\textbf{Leaf optical}\\
                           \textbf{properties}}}
& Chlorophyll       & $\mu$g/cm$^{2}$ & 15    & 80    & LHS & 50 \\
& Water content     & g/cm$^{2}$      & 0.002 & 0.035 & LHS & 50 \\
& Dry matter        & g/cm$^{2}$      & 0.005 & 0.040 & LHS & 50 \\
& Brown pigment     & --              & 0.0   & 0.0   & 0   & 1  \\
& Anthocyanin       & g/cm            & 0.0   & 0.0   & 0   & 1  \\
& Mesophyll structure & --            & 1.5   & 1.5   & 0   & 1  \\

\midrule

\multirow{3}{*}{\makecell[l]{\textbf{Observation}\\
                           \textbf{geometry}}}
& Solar zenith      & $^\circ$ & 15 & 45 & 15 & 3 \\
& Solar azimuth     & $^\circ$ & 90 & 90 & 0  & 1 \\
& View zenith       & $^\circ$ & 0  & 0  & 0  & 1 \\

\bottomrule
\end{tabular}
\end{table}

\paragraph{Geo-foundation models.}
We evaluate three geo-foundation models representing distinct training paradigms:
\begin{itemize}
    \item \textbf{OlmoEarth} \cite{herzog2025olmoearth}, a multimodal, spatio-temporal Earth Observation GFM family. Its encoder is pretrained on three satellite modalities, Sentinel-2, Sentinel-1, and Landsat, with six additional map products used for supervision, including land cover and canopy height. We use the \emph{Base} variant, which reportedly achieves the best overall performance. Its embeddings are 768-dimensional. 
    \item \textbf{Prithvi-v2} \cite{szwarcman2025prithvi}, a GFM pretrained with a masked-autoencoder (MAE) objective on Harmonized Landsat-Sentinel (HLS) imagery. It is therefore trained through self-supervised learning on optical spectral bands.  We use the \emph{300M} variant, whose embeddings are 1,024-dimensional.
    \item \textbf{DINOv3-Sat} \cite{simeoni2025dinov3}, a general vision foundation model fine-tuned on satellite imagery. Unlike the other models considered here, it is trained only on RGB bands and does not take additional spectral bands as input. We use the satellite-pretrained \emph{ViT-L/16} variant, whose embeddings are 1,024-dimensional.
\end{itemize}

These models differ in both input modality and pretraining objective, making them suitable for comparing how representation-learning choices affect concept encoding.

\paragraph{Downstream datasets.}
We use two real-world datasets to evaluate complementary aspects of PiCo. 
The Statewide Land And Trees Survey (SLATS) ground-cover dataset from Australia tests whether sensitivity to physical concepts as identified by the probes can inform hypotheses about a GFM's downstream performance. 
The TreeSatAI tree-species dataset from Germany examines how downstream classes differ along the concept vectors and which concepts influence concept-based predictions. We discuss these in more detail below.

\begin{figure}[htbp]
    \centering
    \includegraphics[width=\columnwidth]{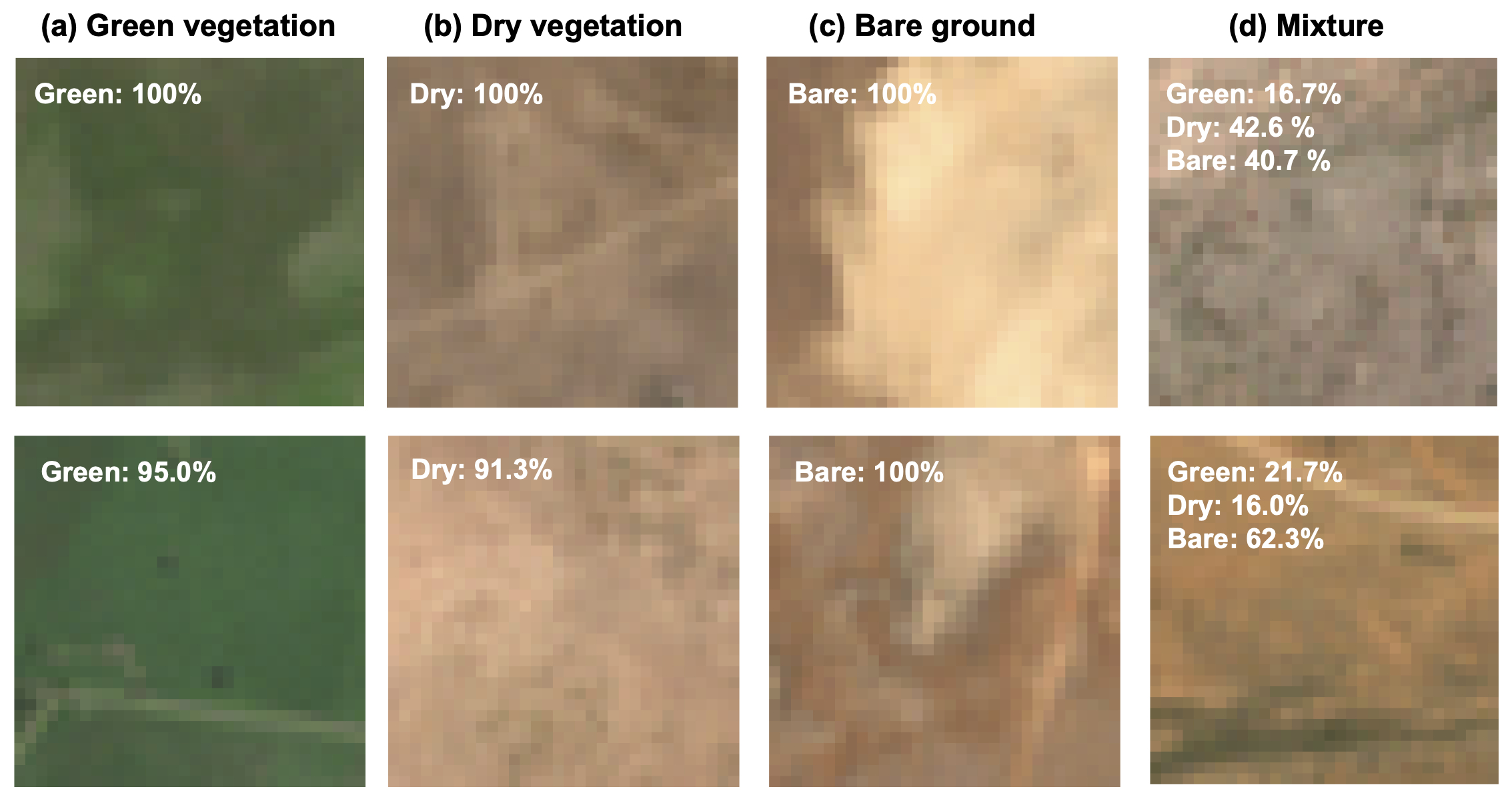}
    \caption{\textbf{The SLATS dataset contains ground cover types with distinct physical properties, enabling hypothesis testing based on model sensitivity to physical concepts.}
    The first three columns show Sentinel-2 RGB patches (320 $\times$ 320 m at 10 m resolution) corresponding to field measurements dominated by a single cover type ($>90\%$), while the last column shows patches containing mixtures of all three types.
    Dry vegetation and bare ground are visually difficult to distinguish despite their physical differences, as dry vegetation contains leaf dry matter, such as proteins and starch, that is absent from bare ground.}
    \label{fig:data_slats_ground_cover_s2.png}
\end{figure}

\textbf{SLATS.}
The SLATS Star Transects--Australian Field Sites dataset (\cref{fig:data_slats_ground_cover_s2.png}) provides field measurements of ground cover at approximately one-hectare sites across Australia, following the protocol of \cite{muir2011factional}.
From the recorded components, we construct three ground-cover targets: \emph{green vegetation}, \emph{dry vegetation and litter} (hereafter \emph{dry vegetation}), and \emph{bare ground}.
The task is to estimate the fractional cover of each type from satellite observations of the site.

We pair 1,472 field measurements from 1,390 Australian sites, including revisits, with Sentinel-2 Level-2A multispectral observations acquired within $\pm15$ days of each measurement.
The measurements span 2017--2026, with most collected during 2017--2019.
For each measurement, we extract a site-centred $320\,\mathrm{m}\times320\,\mathrm{m}$ image window on a $10\,\mathrm{m}$ grid and obtain one GFM embedding by spatially averaging tokens over the central region, corresponding approximately to the field measurement footprint.
Training and test data are separated by measurement site to prevent observations from the same site from appearing in both sets.
A ridge regressor is fitted to the GFM embeddings to predict the field-measured fractional cover of each ground-cover type.

\begin{figure*}[htbp]
    \centering
    \includegraphics[width=\textwidth]{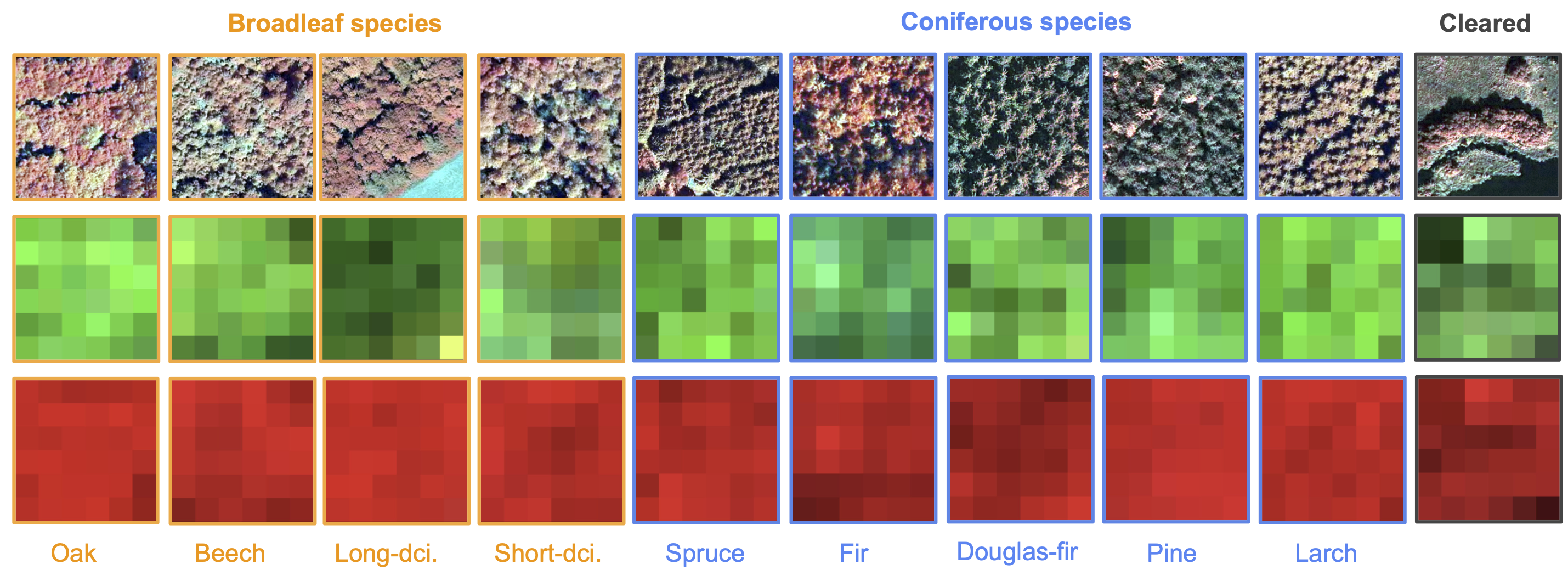}
    \caption{\textbf{TreeSatAI provides a hierarchical tree-species classification dataset with $60\,\mathrm{m}\times60\,\mathrm{m}$ patches dominated by single labels.}
    \textit{We use 10 classes organised into two hierarchical levels. \textbf{First row}: aerial imagery at 0.2 m resolution in false-colour composite (NIR, red, green), where vegetation appears predominantly red due to its strong NIR reflectance. \textbf{Second row}: corresponding Sentinel-2 patches at 10 m resolution in true-colour RGB. \textbf{Third row}: Sentinel-2 patches at 10 m resolution in false-colour composite (NIR, red, green). The aerial imagery is provided for visual reference; the 10 m satellite imagery, commonly used in global Earth observation, is what is evaluated in this study.
    }}
    \label{fig:data_treesatAI_visuals_aerial_s2}
\end{figure*}

\textbf{TreeSatAI.}
TreeSatAI \cite{ahlswede2022treesatai} is a multi-label tree-classification dataset (\cref{fig:data_treesatAI_visuals_aerial_s2}).
The task is to classify forest vegetation labels at different levels of the class hierarchy from Sentinel-2 Level-2A multispectral observations.
Each Sentinel-2 image patch covers $60\,\mathrm{m}\times60\,\mathrm{m}$, corresponding to $6\times6$ pixels at $10\,\mathrm{m}$ resolution.

To limit confounding from mixed-species composition in our analysis, we create a single-label subset by requiring the dominant original label to cover at least $90\%$ of each patch’s labelled area, retaining 28,782 patches.
Following the recommended train/test partition yields 25,916 training samples and 2,866 test samples.
We organise the original labels used in this study into two levels: three coarse forest-cover types (Broadleaf, Conifers, Cleared) and ten finer categories (Oak, Beech, Long-deciduous, Short-deciduous, Spruce, Fir, Douglas-fir, Pine, Larch, Cleared), derived from a genus-to-species taxonomy. 
The \emph{Cleared} class denotes the absence of trees, for example following harvesting, rather than necessarily bare ground.
For fine-grained classification, we evaluate the four broadleaf categories and five coniferous categories separately.
We use a simple $k$NN classifier on the GFM embeddings, following previous GFM classification benchmarks \cite{herzog2025olmoearth}.

%% file: 4_results.tex
\section{Results}
We now present the results from our evaluation, with each section answering a specific research question.
\subsection{Which physical concepts is each GFM sensitive to?}
\label{sec:concept_encoding}
\label{sec:physics_encoding_validation}

We first examine how physical concepts are represented in GFM embeddings from CAMP3D simulations, then assess whether the learned concept vectors remain sensitive to the same physical properties in independently reconstructed canopy scenes.
Since real-world downstream datasets generally lack the biophysical measurements needed to validate concept sensitivity directly, we use 3D RTM simulations from RAMI-V \cite{lanconelli2025fifth}, a radiative transfer benchmark with detailed canopy scenes reconstructed from real-world sites, representing a structural distribution shift relative to CAMP3D scenes. 
This allows us to evaluate transfer under changes in canopy structure while retaining access to ground-truth physical variables.

We first inspect two-dimensional visualisations of embeddings obtained by passing simulated output, with varying leaf-optical and structural properties, through each GFM (details in Section S2 of the Supplementary Material).
In \cref{fig:results_scatter_embedding_scatterplot_rami_validation} (a), dry-matter gradients are visible in the embeddings of OlmoEarth and Prithvi-v2, whereas no comparably clear gradient appears for DINOv3-Sat. 
We observe a similar pattern for water content (Figure S1 of the Supplementary Material). 
These observations suggest a difference in concept encoding according to the input type: DINOv3-Sat uses only RGB bands, whereas OlmoEarth and Prithvi-v2 also receive non-visible bands, including the shortwave infrared (SWIR), which contains absorption features associated with dry matter and water content.

To quantify this insight, we compute the Pearson correlation between each concept projection $\Phi_z(e)$ and its corresponding ground-truth variable $z$ (\cref{fig:results_scatter_embedding_scatterplot_rami_validation} (b)) from RAMI-V.
This measures whether the learned concept vectors remain sensitive to the corresponding physical variables in canopy scenes reconstructed from real-world sites. 
To test the role of input modality as observed above, we additionally evaluate RGB-only variants of OlmoEarth and Prithvi-v2, replacing each non-RGB channel with its corresponding mean value.

\begin{figure*}[htbp]
    \centering
    \includegraphics[width=\textwidth]{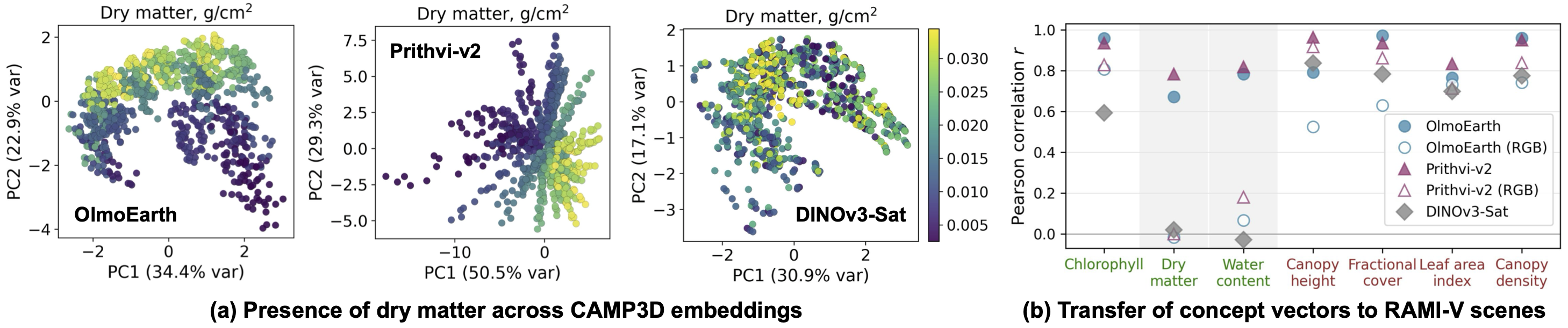}
    \caption{\textbf{Physical concept encoding in CAMP3D embeddings
    and transfer to RAMI-V.}
    \textit{
    (a) The first two principal components of each model's CAMP3D embeddings, coloured by leaf dry-matter content. 
    OlmoEarth and Prithvi-v2 exhibit visible dry-matter gradients, whereas DINOv3-Sat does not show a comparably clear pattern.
    Axis labels report the variance explained by each component. 
    (b) Pearson correlations between CAMP3D-learned concept projections and ground-truth physical variables in RAMI-V.
    RGB ablations replace non-RGB channels with their corresponding mean values. 
    Dry-matter and water-content sensitivity is substantially reduced in the RGB variants and DINOv3-Sat, while chlorophyll and structural concepts remain more accessible.
    }
    }
    \label{fig:results_scatter_embedding_scatterplot_rami_validation}
\end{figure*}

The original OlmoEarth and Prithvi-v2 concept vectors retain sensitivity to dry matter and water content on RAMI-V, with correlations of $r=0.67$--$0.82$. In contrast, dry-matter correlations are approximately zero for DINOv3-Sat and both RGB variants, while water-content correlations are weak ($r=-0.03$--$0.18$).  
Chlorophyll remains accessible across all models and variants, although its correlation is lower for DINOv3-Sat ($r=0.59$). 
This is consistent with chlorophyll absorption in the visible red band. 
Structural concept vectors also retain positive correlations across all models and variants ($r=0.52$--$0.97$), although RGB restriction reduces transfer performance for several concepts.

\begin{figure*}[htbp]
    \centering
    \includegraphics[width=\textwidth]{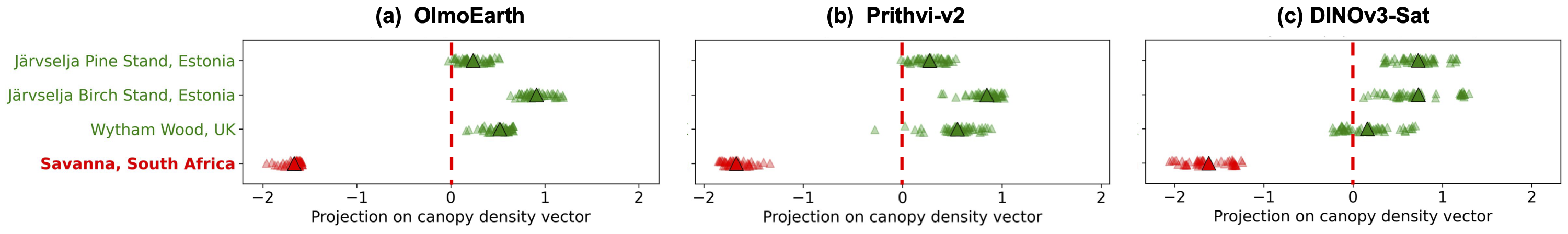}
    \caption{\textbf{The African savanna projects to the low-density end of the concept vector for canopy-density.}
    \textit{
    Projections onto CAMP3D-learned concept vectors are shown for (a) OlmoEarth, (b) Prithvi-v2, and (c) DINOv3-Sat.
    Each RAMI-V scene contains 50 samples with varying leaf-optical properties and fixed canopy structure. 
    Small markers denote individual samples; large outlined markers denote scene means.
    Projections are standardised separately for each model using the mean and standard deviation across the RAMI-V samples shown; the dashed line marks the mean projection.
    All three models place the sparse savanna below the forest scenes. DINOv3-Sat shows greater within-scene dispersion, despite the canopy structure being fixed for each scene. 
    }
    }   \label{fig:results_scatter_density_projection_olmo_prithvi_dino}
\end{figure*}

\cref{fig:results_scatter_density_projection_olmo_prithvi_dino} provides a visual example of a concept vector (canopy density) that is expressed in all three models.
Each RAMI-V scene contains 50 samples generated using the same Latin hypercube sampling of the three leaf-optical variables.
For all three models, the African savanna projects towards the low-density end of the concept vector, consistent with its sparse canopy structure. 
However, DINOv3-Sat exhibits a wider within-scene spread, which suggests that its canopy-density projection is more sensitive to variation in leaf optics than those of OlmoEarth and Prithvi-v2. 

\subsection{Does a GFM’s physical sensitivity determine its downstream performance?} 

The RAMI-V validation shows that DINOv3-Sat and the RGB variants of OlmoEarth and Prithvi-v2 have limited sensitivity to leaf dry matter and water content. 
We use domain knowledge to relate these sensitivity differences to the SLATS ground-cover estimation task, which uses satellite imagery to estimate field-measured fractions of green vegetation, dry vegetation, and bare ground.
These three ground cover types differ in their physical properties. 
Green vegetation contains chlorophyll, whereas dry vegetation includes senescent plant material and detached litter, in which chlorophyll is depleted but dry matter, such as cellulose and lignin, remains.
Bare ground includes exposed soil and rock, which lack both leaf chlorophyll and dry-matter properties.
We therefore hypothesise that limited dry-matter sensitivity will be associated with poorer fractional-cover estimation for dry vegetation and bare ground than for green vegetation, for which chlorophyll sensitivity is comparatively retained in all models.

\begin{figure}[htbp]
    \centering
    \includegraphics[width=\columnwidth]{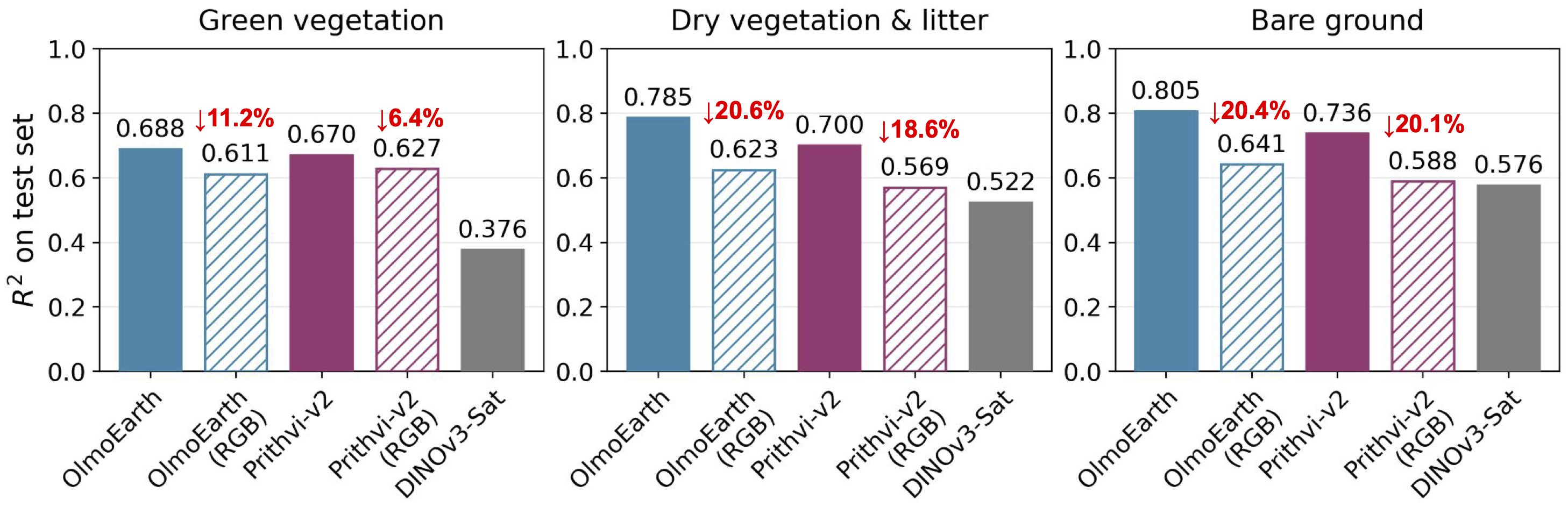}
    \caption{\textbf{RGB restriction disproportionately degrades fractional-cover estimation of dry vegetation and bare ground.}
    \textit{
    Bars show $R^2$ for Ridge regressors fitted to pre-computed GFM embeddings. Solid bars denote native inputs, and hatched bars denote RGB ablations in which non-RGB channels are replaced by their corresponding mean values.
    Red annotations indicate relative reductions in $R^2$ between the each GFM and its RGB variant.
    DINOv3-Sat uses RGB inputs only, so no RGB ablation is applicable.
    The larger reductions for dry vegetation and bare ground are consistent with the failure mode anticipated from limited dry-matter sensitivity.
    }
    }
    \label{fig:results_bar_fcover_r2}
\end{figure}

We test this hypothesis using the SLATS ground-cover dataset, and the results support it (\cref{fig:results_bar_fcover_r2}). 
DINOv3-Sat performs worse than the original OlmoEarth and Prithvi-v2 on dry vegetation and bare ground. 
Additionally, the within-model RGB ablations produce larger relative reductions in $R^2$ for these two targets than for green vegetation. 
For OlmoEarth and Prithvi-v2, respectively, $R^2$ decreases by 20.6\% and 18.6\% for dry vegetation, and by 20.4\% and 20.1\% for bare ground, compared with 11.2\% and 6.4\% for green vegetation. 

Our findings therefore challenge DINOv3's claim that a generic RGB-based training recipe can be applied to Earth observation ``out of the box'' \cite{simeoni2025dinov3}. 
More broadly, strong performance on existing benchmarks does not guarantee that a GFM captures the physical concepts relevant to a particular downstream task. 
To our knowledge, PiCo provides the first demonstration that task-specific performance can be anticipated solely by probing the physical concepts encoded in GFM embedding space, even \emph{before} training or evaluating a downstream predictor.

\subsection{How do concept representations differ across downstream classes and explain predictions?}
\label{sec:decision_pattern}

We next apply the concept representation $\Phi(\mathbf{e})$, as defined in \cref{eq:concept_projected_representation}, to downstream forest-type and tree-species classification using the TreeSatAI dataset.
We examine two aspects of $\Phi(\mathbf{e})$: how projections of different downstream classes are distributed along physical concept vectors, and which concepts contribute most to their classification.
We focus on OlmoEarth and Prithvi-v2, which we have found to be sensitive to all seven concepts, enabling a consistent comparison. More details can be found in Section S3 of the Supplementary Material.

\begin{figure}[htbp]
    \centering
    \includegraphics[width=\columnwidth]{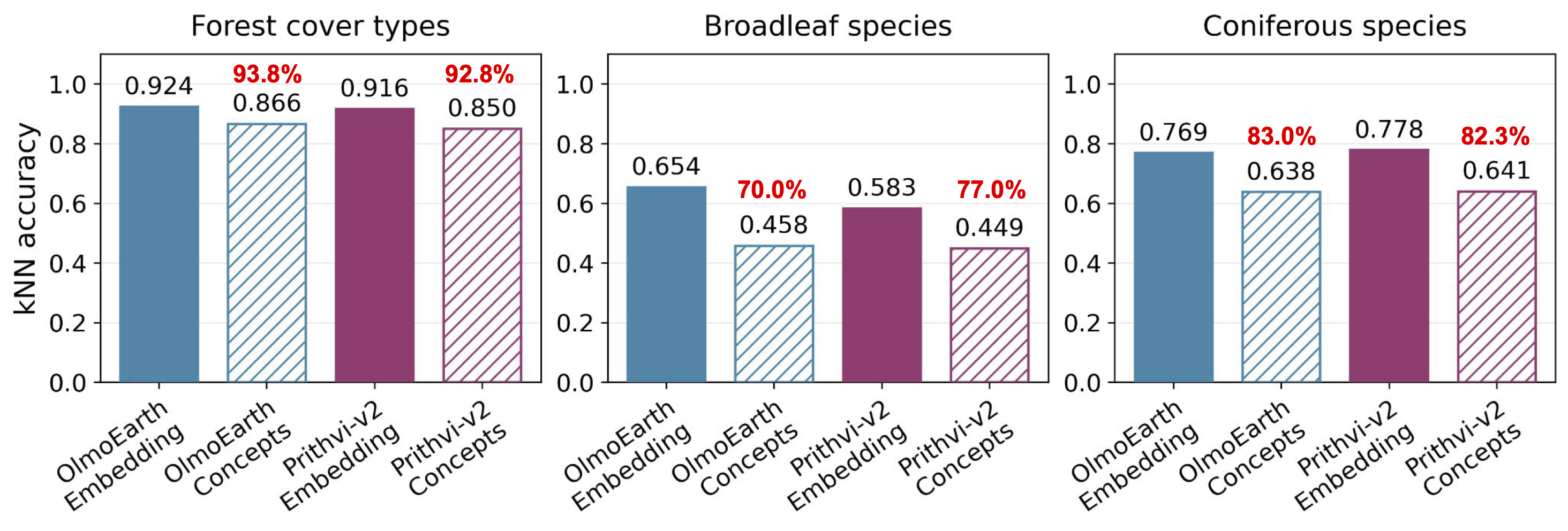}
    \caption{\textbf{Predictive sufficiency of physical concept representations on TreeSatAI}
    \textit{
    Solid and hatched bars show $k$NN classification accuracy using full embeddings and concept representations, respectively.
    Red annotations report retained accuracy when comparing the concept prediction with the embedding prediction of each model.
    Concept representations retain 92.8--93.8\% of full-embedding accuracy for forest cover types and 70.0--83.0\% for species-level classification.
    }
    }
    \label{fig:results_bar_treesat_accuracy}
\end{figure}

Following previous work on explanations based on low-dimensional representations \cite{yeh2020completeness,zhou2018interpretable}, we first assess the predictive sufficiency of the concept representation $\Phi(\mathbf{e})$ as a surrogate for the full high-dimensional embedding $\mathbf{e}$.
\Cref{fig:results_bar_treesat_accuracy} compares $k$NN classification accuracy using the full embeddings and their seven-dimensional concept representations.
The concept representations retain 92.8--93.8\% of the full-embedding accuracy for forest cover types and 70.0--83.0\% for species-level classification.
The reduction in accuracy, especially for species-level classification, reflects the trade-off between classification accuracy and interpretability when using only low-dimensional concept representations.
In practice, it may be preferable to retain the original embeddings for classification alongside the concept representations for interpretation.
This preserves classification accuracy, although the resulting explanations become approximate.

\begin{figure*}[htbp]
    \centering
    \includegraphics[width=\textwidth]{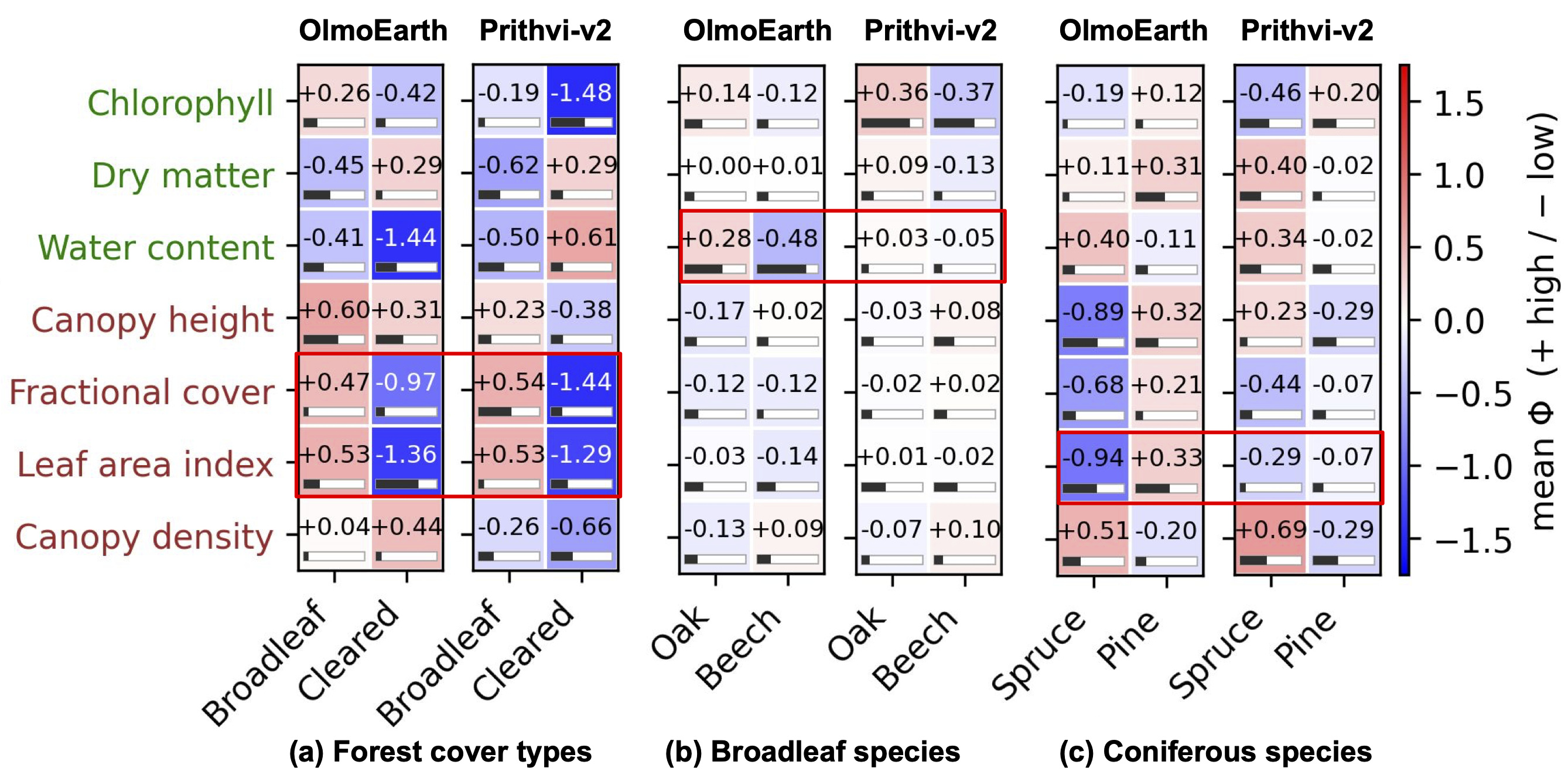}
    \caption{\textbf{Concept projections reveal physical contrasts between classes, while concept importance differs between models.}
    \textit{
    Colours and numbers show signed class-wise mean concept projections for OlmoEarth and Prithvi-v2.
    Inset bars show normalized SHAP importance, with a full bar corresponding to a value of 0.5. Importance scores sum to one across the seven concepts for each class.
    Classes highlighted with red boxes illustrate (a) structural contrasts between broadleaf forest and cleared areas, (b) leaf-optical contrasts between oak and beech, and (c) the leaf area index interpretation examined for spruce and pine.
    Attributions are computed within each full classification task, rather than only between the displayed class pairs.
    }
    } 
    \label{fig:results_heatmap_projection_shap_olmo_prithvi}
\end{figure*}

We then analyse the individual concept projections $\Phi_k(\mathbf{e})$ 
and assess their importance for classification across different levels of the TreeSatAI class hierarchy (\cref{fig:data_treesatAI_visuals_aerial_s2}), as quantified by the score function defined in \cref{eq:normalised_concept_importance}.
At the top level, the TreeSatAI class hierarchy distinguishes forest cover types, including \emph{broadleaf}, \emph{conifer}, and \emph{cleared}; at the next level, it distinguishes species within the broadleaf and conifer groups.
\Cref{fig:results_heatmap_projection_shap_olmo_prithvi} presents two complementary quantities. Signed class-wise mean projections describe where classes lie along each concept vector, allowing comparison with ecological expectations.
Normalised SHAP importance scores describe the relative contribution of each concept to the concept-based classifier.
We refer to the concept with the highest importance score for a class as its \emph{dominant concept}.

For forest cover types, broadleaf forest and cleared areas have an intuitive structural contrast: cleared areas contain little or no remaining tree cover.
\cref{fig:results_heatmap_projection_shap_olmo_prithvi} shows the standardised concept projections, averaged across test samples for each class (colours and numbers), along with their normalised importance scores for classification as defined in \cref{eq:normalised_concept_importance} (inset bars).
Both models reflect this distinction in their fractional-cover and leaf-area-index projections (\cref{fig:results_heatmap_projection_shap_olmo_prithvi} (a)).
For OlmoEarth, the cleared class has strongly negative mean projections on fractional cover ($-0.97$) and leaf area index ($-1.36$), consistent with reduced canopy cover.
Leaf area index is also its dominant concept for this class, with an importance score of $0.35$. 
Prithvi-v2 shows similarly negative projections on fractional cover ($-1.44$) and leaf area index ($-1.29$), but a stronger negative chlorophyll projection ($-1.48$, compared with $-0.42$ for OlmoEarth).
Its dominant concept is chlorophyll, with an importance score of $0.28$, compared with $0.14$ for leaf area index.

For broadleaf species, oak and beech illustrate a contrast associated with leaf-optical properties (\cref{fig:results_heatmap_projection_shap_olmo_prithvi} (b)).
These co-occurring deciduous trees differ in transpiration patterns and water-use strategies \cite{fabiani2022transpiration}, motivating an examination of their water-content projections.
OlmoEarth places oak higher than beech on this axis ($+0.28$ versus $-0.48$), and water content is the dominant concept for both classes, with importance scores of $0.31$ and $0.40$, respectively. 
Prithvi-v2 shows the same ordering but a much weaker water-content contrast ($+0.03$ versus $-0.05$).
Instead, it exhibits a stronger chlorophyll contrast ($+0.36$ versus $-0.37$), and chlorophyll is the dominant concept for both classes.
The dominant concepts also depend on class hierarchy: structural contrasts are prominent between broadleaf forest and cleared areas, whereas leaf-optical contrasts can be more prominent for fine-grained classifications in the oak--beech example. 

Together, these examples show that both models express physically meaningful contrasts along the concept vectors, but could differ in the strength of these contrasts and in which concepts ultimately dominate classification.

\subsection{Can concept projections mislead?} 
\label{sec:spectral_observability}

Concept projections can mislead: for example, a physical characteristic, leaf area index, could be misinterpreted based on concept projections alone (\cref{fig:results_heatmap_projection_shap_olmo_prithvi} (c)).
Field measurements commonly associate spruce stands with a higher leaf area index (leaf area per unit ground area) than pine stands \cite{goude2019comparing}. 
Their contrasting canopy appearances in the TreeSatAI dataset are
visualised in \cref{fig:data_treesatAI_visuals_aerial_s2}.
Leaf area index is among the most important concepts for OlmoEarth when classifying spruce and pine, with importance scores of $0.27$ and $0.28$, respectively.
However, their relative positions along the leaf-area-index concept vector do not follow the expected ordering: spruce has the lowest projection in both models ($-0.94$ for OlmoEarth and $-0.29$ for Prithvi-v2), despite its typically dense foliage.

To investigate this inconsistency, we analyze the raw near-infrared (NIR) reflectance across test samples from the conifer classes.
NIR reflectance is widely used in vegetation remote sensing because leaves absorb relatively little NIR radiation and instead reflect and transmit much of it.
A denser canopy with higher leaf area index can therefore increase multiple scattering and, in turn, the recorded NIR reflectance \cite{ollinger2011sources}.
This relationship is modelled in the CAMP3D scenes, and the learned leaf-area-index concept vector captures the associated shifts in the embeddings computed from these observations.

As shown in \cref{tab:valiation_nir_reflectance}, spruce also has the lowest mean NIR reflectance (17.9\%) among the conifer classes.
This could help explain its low projection along the leaf-area-index concept vector, which is learned from CAMP3D scenes where low projections along this vector are associated with low NIR reflectance.
However, this may involve confounding: previous studies suggest that crown self-shadowing, rather than leaf area index, can dominate the NIR response in coniferous forests \cite{kane2008interpretation}.
As crowns become denser, self-shadowing can outweigh the increase in NIR reflectance associated with higher leaf area index.
In particular, \cite{williams1991comparison} found that self-shadowing in dense spruce crowns led to lower NIR reflectance than in other conifers, including red pine and white pine.
Interpreting spruce's low projection without this context could incorrectly suggest low leaf area index.
Thus, the physical interpretation of concept representations may require species-specific context when other factors, such as self-shadowing, confound the spectral observability of physical states.

\begin{table}[htbp]
    \centering
    \caption{\textbf{Spruce has the lowest mean near-infrared reflectance (\%) among the conifer classes.}
    \textit{
    Reflectance is averaged over test samples for each class. The low spruce reflectance does not necessarily indicate low leaf area index; it could instead reflect crown self-shadowing due to its typically dense foliage, confounding interpretation.
    }
    }
    \label{tab:valiation_nir_reflectance}
    \resizebox{\columnwidth}{!}{
    \begin{tabular}{lccccc}
        \toprule
        & \textbf{Spruce} & \textbf{Fir} & \textbf{Douglas fir}
        & \textbf{Pine} & \textbf{Larch} \\
        \midrule
        Number of samples & 384 & 65 & 153 & 564 & 147 \\
        Mean NIR reflectance & \textbf{17.9} & 24.4 & 23.5 & 23.3 & 24.7 \\
        \bottomrule
    \end{tabular}
    }
\end{table}

%% file: 5_conclusion.tex
\section{Conclusion}
We introduced PiCo --- Physics-Informed Concept Probing --- a post-hoc framework for explaining predictions from GFM embeddings through physical concepts.
PiCo combines 3D radiative transfer modelling with a game engine to generate large-scale synthetic multispectral satellite imagery with controlled physical variations, enabling the identification of concept vectors in GFM embeddings. 
We applied PiCo to three GFMs trained on distinct modalities and assessed their explainability through seven core physical concepts covering leaf optics and forest vegetation structure.

Our experiments show that concept vectors learned from forest scenes built using game-engine assets transfer to independently reconstructed canopies from real-world sites, retaining sensitivity to the corresponding physical concepts.
The sensitivities of GFMs to these physical concepts help formulate and test domain-informed hypotheses about downstream performance, such as whether limited sensitivity to leaf dry matter hinders discrimination between dry vegetation and bare ground.
Using only seven physical concepts (i.e. only seven dimensions), the concept representations retain 70.0--93.8\% of full-embedding accuracy across the class hierarchy in a tree species classification dataset.

Our analysis of concept importance in downstream predictions reveals that models can capture similar physical contrasts between classes, such as structural differences between broadleaf forest and cleared areas and water-related differences between oak and beech, yet rely on different concepts for classification. This lays the groundwork for future research to align model decisions with domain knowledge.
These concept-based interpretations also need to consider how observable the underlying physical states are, since other factors can confound their spectral signatures and embedding responses.

PiCo could extend to scientific foundation models when mechanistic observation models enable controlled simulations to identify physical concepts in learned representations.
Linking physical simulations to data-driven representations through PiCo therefore enables hypothesis testing during model development and helps scientists assess the physical basis of model decisions.